\documentclass[10pt,twocolumn,letterpaper]{article}

\usepackage[pagenumbers]{iccv} 

\usepackage{microtype}
\usepackage{graphicx}
\usepackage{booktabs} 
\usepackage{amsmath}
\usepackage{caption}
\usepackage{dblfloatfix}

\usepackage{fancyhdr}
\definecolor{iccvblue}{rgb}{0.21,0.49,0.74}
\usepackage[pagebackref,breaklinks,colorlinks,allcolors=iccvblue]{hyperref}

\def\paperID{15688} 
\def\confName{ICCV}
\def\confYear{2025}

\newcommand\nonumfootnote[1]{%
\begingroup%
    \renewcommand\thefootnote{}\footnote{\hspace{-3.7pt}#1}%
    \addtocounter{footnote}{-1}%
\endgroup%
}

\title{StreamDiffusion: A Pipeline-level Solution for Real-time Interactive Generation}

\author{Akio Kodaira$^{1*}$ \hspace{0.02cm}
Chenfeng Xu$^{1,*}$ \hspace{0.02cm}
Toshiki Hazama$^{1,*}$ \hspace{0.02cm}
Takanori Yoshimoto$^2$ \hspace{0.02cm}
Kohei Ohno$^3$ \hspace{0.02cm}\\
Shogo Mitsuhori$^4$ \hspace{0.02cm}
Soichi Sugano$^5$ \hspace{0.02cm}
Hanying Cho$^6$ \hspace{0.02cm}
Zhijian Kiu$^7$ \hspace{0.02cm}
Masayoshi Tomizuka$^1$ \hspace{0.02cm}
Kurt Keutzer$^1$\\
\\
$^1$UC Berkeley \hspace{0.02cm}
$^2$University of Tsukuba \hspace{0.02cm}
$^3$International Christian University\\
$^4$Toyo University \hspace{0.02cm}
$^5$Tokyo Institute of Technology \hspace{0.02cm}
$^6$Tohoku University \hspace{0.02cm}
$^7$MIT \vspace{0.1cm}\\
\url{https://github.com/cumulo-autumn/StreamDiffusion} \\
\{akio.kodaira, xuchenfeng\}@berkeley.edu
}

\begin{document}
\twocolumn[{
\maketitle
\begin{center}
    \vspace{-6mm}
	\centering
    \small
	\captionsetup{type=figure}
 \includegraphics[width=\textwidth]{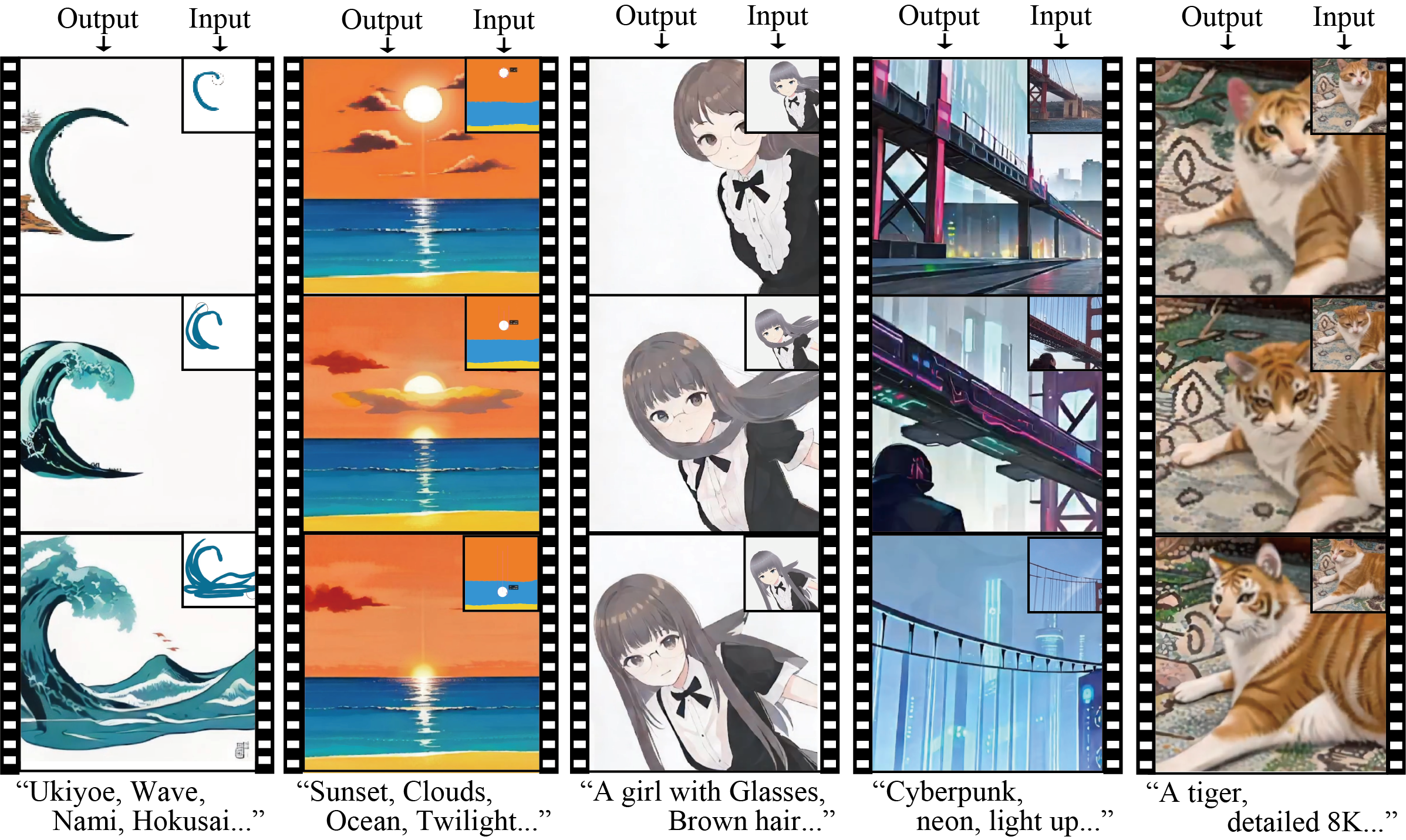}
    \captionof{figure}{StreamDiT-4B: video generation can be streaming and real-time.}
    \vspace{1mm}
    \label{fig:teaser}
\end{center}
}]
\maketitle
\begin{abstract}
\nonumfootnote{\noindent $^*$ denotes equal contribution}
\nonumfootnote{This work was done when Toshiki was a remote intern at UC Berkeley}
We introduce StreamDiffusion, a real-time diffusion pipeline designed for streaming image generation. Existing diffusion models are adept at creating images from text or image prompts, yet they often fall short in real-time interaction. This limitation becomes particularly evident in scenarios involving continuous input, such as augmented/virtual reality, video game graphics rendering, live video streaming, and broadcasting, where high throughput is imperative. 
StreamDiffusion tackles this challenge through a novel pipeline-level system design. It employs unique strategies like batching the denoising process (Stream Batch), residual classifier-free guidance (R-CFG), and stochastic similarity filtering (SSF). Additionally, it seamlessly integrates advanced acceleration technologies for maximum efficiency.
Specifically, Stream Batch reformulates the denoising process by eliminating the traditional wait-and-execute approach and utilizing a batching denoising approach, facilitating fluid and high-throughput streams. This results in 1.5x higher throughput compared to the conventional sequential denoising approach. R-CFG significantly addresses inefficiencies caused by repetitive computations during denoising. It optimizes the process to require minimal or no additional computations, leading to speed improvements of up to 2.05x compared to previous classifier-free methods. Besides, our stochastic similarity filtering dramatically lowers GPU activation frequency by halting computations for static image flows, achieving a remarkable reduction in computational consumption—2.39 times on an RTX 3060 GPU and 1.99 times on an RTX 4090 GPU, respectively. The synergy of our proposed strategies with established acceleration technologies enables image generation to reach speeds of up to 91.07 fps on a single RTX 4090 GPU, outperforming the throughput of AutoPipeline, developed by Diffusers, by more than 59.56x. 
\end{abstract}

\section{Introduction}
\label{sec:intro}

Recently, there has been a growing trend in the commercialization of diffusion models \cite{rombach2022high,Saharia2022,BetkerImprovingIG,sora} for applications within the entertainment industry such as Metaverse, online video streaming, broadcasting, and even the robotic field \cite{chi2023diffusion}. A pertinent example is the use of diffusion models to create virtual YouTubers. These digital personas should be capable of reacting in a fluid and responsive manner to user input. These areas require diffusion pipelines that offer high throughput and low latency to ensure the efficient interactive streaming generation. 

To advance the efficiency, current efforts primarily focus on reducing the number of denoising steps, such as decreasing from 50 denoise steps to just a few \cite{luo2023latent,luo2023lcm} or even one \cite{yin2023onestep,liu2023insta}. The strategy includes distilling the multi-step diffusion models into a few steps \cite{sauer2023adversarial,song2023consistency} or re-framing the diffusion process with neural Ordinary Differential Equations (ODE) \cite{lu2022dpm,lu2022dpm+}. 
Quantization has also been applied to diffusion models \cite{li2023qdiffusion,huang2023tfmq} to improve efficiency. These methods share the common goal of approximating either the diffusion process itself or the original model weights to achieve efficiency gains.
In this paper, we aim at an orthogonal direction and introduce StreamDiffusion, a pipeline-level solution that enables streaming image generation with high throughput. We highlight that existing model design efforts can still be integrated with our pipeline. Our approach enables the use of N-step denoising diffusion models while keeping high throughput and offers users more flexibility in choosing their preferred models.

\begin{figure*}[ht]
  \centering
   \includegraphics[width=\linewidth]{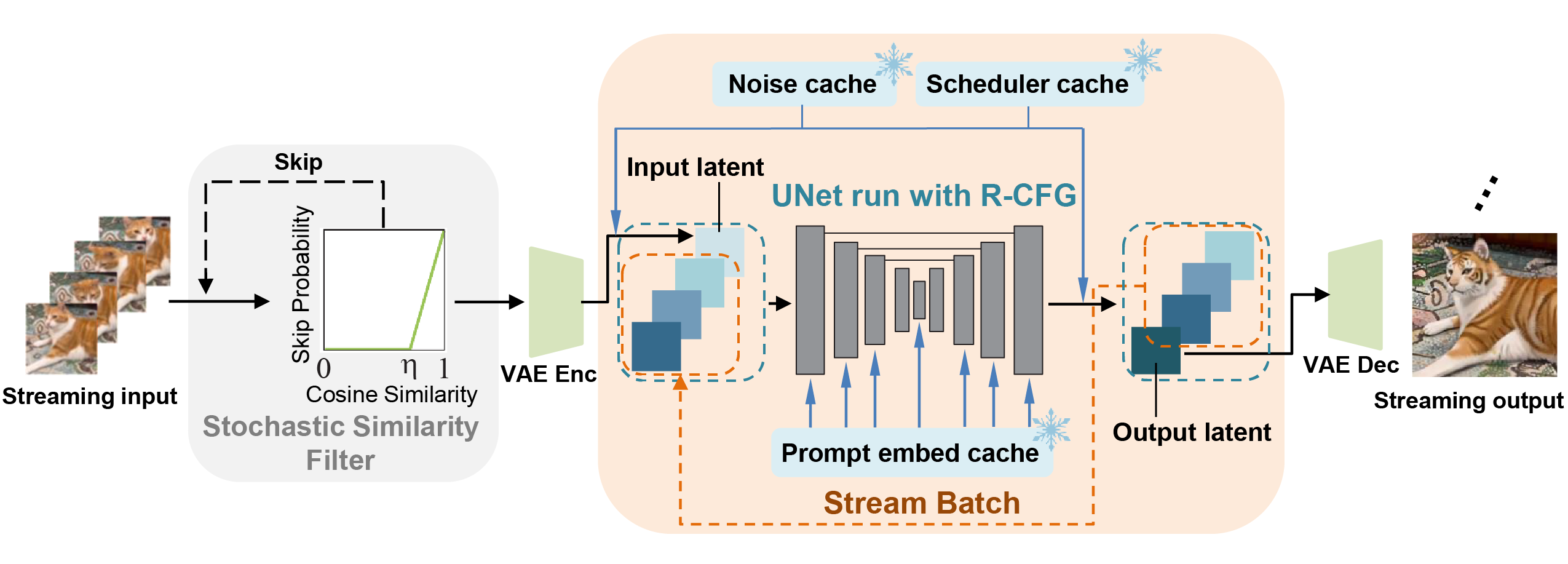}
\caption{The overview of StreamDiffusion. StreamDiffusion combines several key components: (1) Stream Batch efficiently processes the denoising steps in batches. (2) Residual Classifier-Free Guidance approximates the negative condition term to reduces the unnecessary repetitive calculations in the UNet. (3) Stochastic Similarity Filter controls the pass of the image stream by calculating the similarity between frames to eliminate the redundant hit onto GPUs. Furthermore, StreamDiffusion leverages techniques like input-output queues for smooth data flow, cache management for faster processing through pre-calculated embedding, and a tiny VAE model to further contribute to overall efficiency. This synergistic combination allows StreamDiffusion to generate high-quality images at high throughputs while consuming minimal energy. }
\vspace{-5mm}
\label{fig:overview}
\end{figure*}

Specifically, StreamDiffusion seamlessly incorporates a suite of novel strategies. Among these, we propose a simple yet novel approach termed \textbf{Stream Batch}. This method differs from the traditional sequential denoising mode, instead of batching the denoising steps. This subtle modification enhances efficiency without sacrificing the quality of image generation. We highlight Stream Batch enables a new capability of generating images conditioned on future frames in a streaming mode, which is something impossible in previous works. Via injecting the future frames, Stream Batch significantly improves the temporal consistency with few additional overhead.
Furthermore, the key novelty of Stream Batch lies not only in its GPU parallelization; rather, it serves as a practical realization of a broader Stream Denoising framework. By denoising inputs at diagonally-offset timesteps within a streaming queue structure, diffusion models naturally achieve continuous, autoregressive generation—emitting one output frame per each newly sampled frame—while benefiting from parallel computation. Crucially, this design generalizes directly to sequential tasks such as video, audio, or robotic action-sequence generation, enabling interactive, unbounded-length synthesis.

Besides, we point out that it is time-consuming for existing diffusion pipelines to use classifier-free guidance for emphasizing the prompts during generation, due to the repetitive and redundant computations for negative conditions. To address this issue, we introduce an innovative approach termed as \textbf{residual classifier-free guidance (R-CFG)}. R-CFG approximates the negative condition with a virtual residual noise, which allows us to calculate the negative condition noise only during the initial step of the process. We also indicate that using the original input image latent as the residual term effectively generates results that diverge from the original input image according to the magnitude of the guidance scale, which is a special case of our R-CFG and does not require any computations for the negative condition term. 

Furthermore, in real applications such as virtual youtuber and AR/VR cases, maintaining the diffusion models always in an active mode is energy-consuming as it keeps hitting GPU. To reduce the energy, we further apply a \textbf{stochastic similarity filtering (SSF)} strategy. In the pipeline, we compute the similarities between continuous inputs and determine whether the diffusion model should process the images based on the probability of an activated similarity. This enables both energy efficiency and visual fluency. In order to further improve the efficiency to cater to the real applications, we apply simple yet effective engineering implementations such as Input-Output Queue (IO-Queue), pre-computing caching, and TensorRT. The overview of the StreamDiffusion pipeline is shown in Fig. \ref{fig:overview}.

Experiments demonstrate that our proposed StreamDiffusion can achieve up to 91.07fps for image generation on one RTX4090 GPU, surpassing the diffusion Autopipeline from Diffusers \cite{diffusers} team by up to 59.6x. Besides, our stochastic similarity filtering strategy significantly reduces the GPU power usage by 2.39x on one RTX 3090GPU and by 1.99x on one RTX 4090GPU. Our proposed StreamDiffusion is a new diffusion pipeline that is not only efficient but also energy-saving.

\section{Related work}
\label{sec:intro}
\paragraph{\textbf{Efficient Diffusion Models}}
Diffusion models \cite{song2020score,ho2020denoising,rombach2022high,Peebles2022DiT} have sparked considerable interest in the commercial sector due to their high-quality image/video generation capabilities. These models have been progressively adapted for various applications, including text-to-image generation \cite{rombach2021highresolution,ramesh2022hierarchical,avrahami2023blendedlatent}, image editing \cite{Avrahami_2022_CVPR,ruiz2022dreambooth}, video generation \cite{blattmann2023stable,blattmann2023align} and even perception \cite{xu20233difftection,li2023grounded,khani2024slime}. However, diffusion models are currently limited by their slow speed in generating outputs.
In response to this challenge, a variety of strategies have been proposed. One of the mainstreams is to approximate the SDE-based diffusion process \cite{song2020score,song2021denoising} through an ordinary differentiable equation (ODE) framework. For example, DPM and DPM++ \cite{lu2022dpm,lu2022dpm+} introduce ODE-based samplers, which significantly reduce the hundreds of denoising steps to between 15 and 20. Building upon the ODE formulation, InstaFlow \cite{liu2023insta} advances the reduction of denoising steps to a single instance through the novel strategy of rectified flow \cite{liu2022flow}, while achieving performance close to that of Stable Diffusion \cite{rombach2022high}. Additionally, distillation from pre-trained diffusion models has also been explored as a method to facilitate few-step denoising. For instance, the consistency model \cite{song2023consistency} leverages the principle of self-consistency between noise at different denoising steps and uses pre-trained diffusion models \cite{rombach2022high} to guide the learning of a few-step denoising model, thereby enabling the generation of images within a minimal number of steps. In a notable extension of this concept, LCM \cite{luo2023lcm,luo2023lcmlora} applies the idea to the latent space rather than the pixel space. The use of distillation methods \cite{sauer2023adversarial,salimans2022progressive,yin2023onestep} to enhance the efficiency of the original Stable Diffusion model presents promising results.
Besides improving efficiency through the lens of reducing denoising steps, quantization methods \cite{li2023qdiffusion,huang2023tfmq} are proposed to make the model run in the regime of low float points, albeit with the potential trade-off of fidelity and efficiency.
Moreover, parallel sampling \cite{shih2024parallel} tries to utilize the approximate-parallel denoising strategy to improve the latency of the diffusion model. We emphasize that our work focuses on enhancing throughput and our work is orthogonal to \cite{shih2024parallel}. Notably, our method is optimized for single-GPU, which are common among users. In contrast, the parallel sampling method reduces throughput on a single GPU. 

Our proposed StreamDiffusion is significantly different from the approaches mentioned previously. While earlier methods primarily focus on the low latency of their individual model designs, our approach takes a different route. We introduce a \textit{pipeline-level} solution specifically tailored for high throughput. Our pipeline can seamlessly integrate the low-latency diffusion models discussed above. Our proposed Stream Batch, residual classifier-free guidance, and the integration of other efficiency-enhancement methods focus on improving the efficiency of the whole pipeline instead of a single diffusion model.

\begin{figure}[!t]
  \centering
   \includegraphics[width=1.0\linewidth]{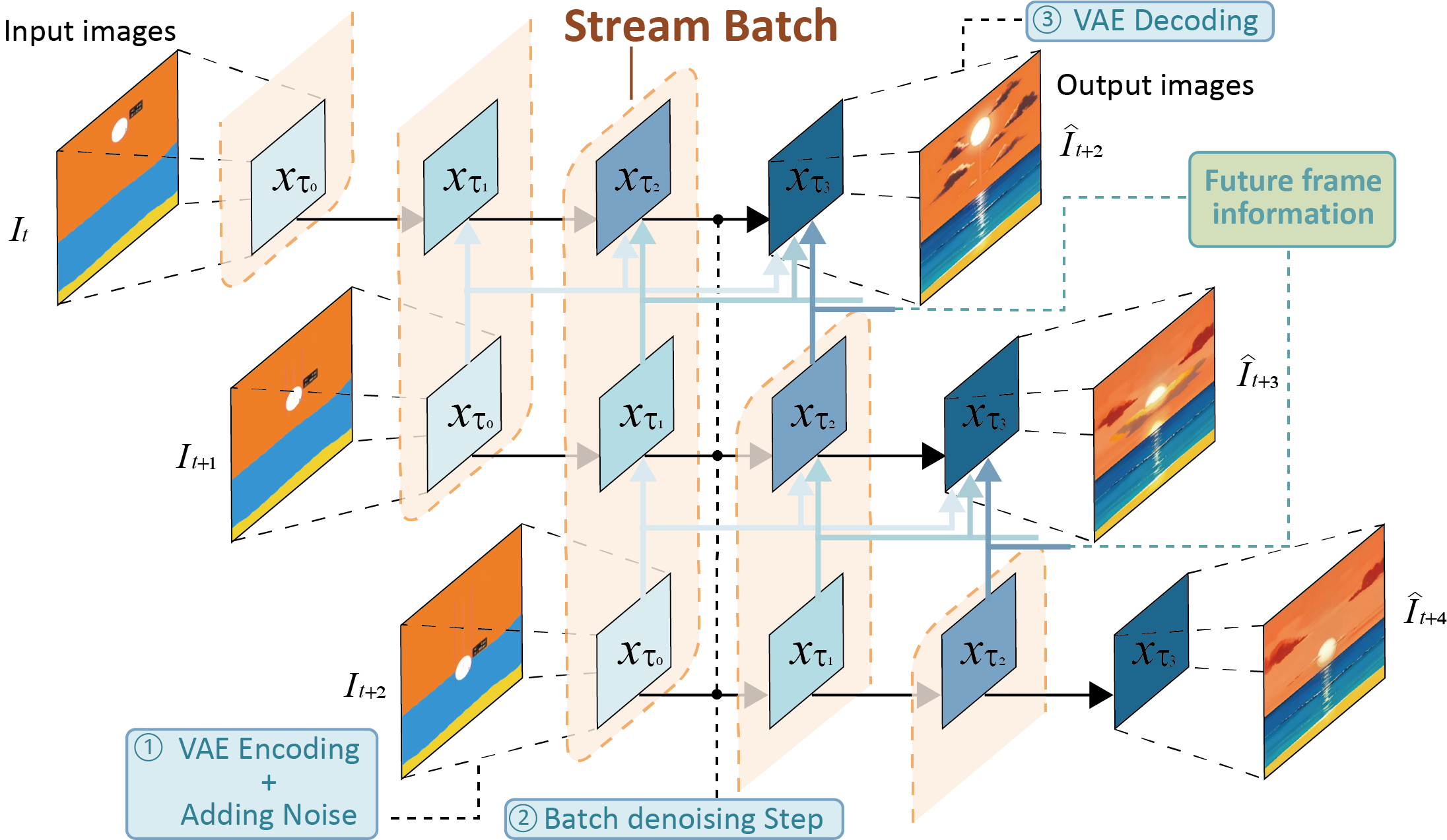}
\caption{The concept of Stream Batch. In our approach, instead of waiting for a single image to be fully denoised before processing the next input image, we accept the next input image after each denoising step. This creates a denoising batch where the denoising steps are staggered for each image. By concatenating these staggered denoising steps into a batch, we can efficiently process continuous inputs using a U-Net for batch processing. The input image encoded at timestep $t$ is generated and decoded at timestep $t+n$, where $n$ is the number of the denoising steps.}
\label{fig:batch_concept}
\end{figure}

\paragraph{\textbf{Classifier-free Guidance}}
Classifier-free guidance \cite{ho2022classifier} is widely used for conditional generation due to the simplicity, efficiency, and stability compared to classifier-guidance \cite{dhariwal2021diffusion}. It leverages negative prompts \cite{crowson2022vqgan,du2020compositional,rombach2022high} and essentially operates the vector arithmetic shift
in latent space, \textit{i.e.,} we take a step of size (usually set by the guidance scale) away from the unconditional vector or negatively conditioned vector in the
direction toward the conditioning manifold \cite{ho2022classifier}. In the practical implementation, the classifier-free guidance is conducted by sharing the same UNet for both the conditional term and unconditional (or negative) term and subtracting the effect of the unconditional term from the conditioned one. We point out that the way of multiple denoising processes leads to unnecessary computations for the unconditional (or negative) term. To get rid of these redundant computations, we propose a novel residual-classifier-free guidance, termed as \textbf{R-CFG}, which approximates the conditional noise prediction with only requiring one or even zero-time computation for the negatively conditioned noise prediction for the UNet. We note that \textit{R-CFG} is especially designed for the SDEdit method \cite{meng2022sdedit}, as we mainly focus on the applications of translating streaming image flow. Our proposed R-CFG significantly improves the latency of the conditional image-to-image generation.

\section{StreamDiffusion}
\label{sec:intro}

StreamDiffusion is a new diffusion pipeline aiming for high throughput. It comprises three key components: Stream Batch strategy, Residual Classifier-Free Guidance (R-CFG), and Stochastic Similarity Filter. Besides, we also incorporate other acceleration methods like a novel input-output queue designed by us, the pre-computation procedure, the tiny-autoencoder, and model acceleration tools such as TensorRT. We elaborate on the details below.

\begin{figure*}[!t]
  \centering
   \includegraphics[width=.95\linewidth]{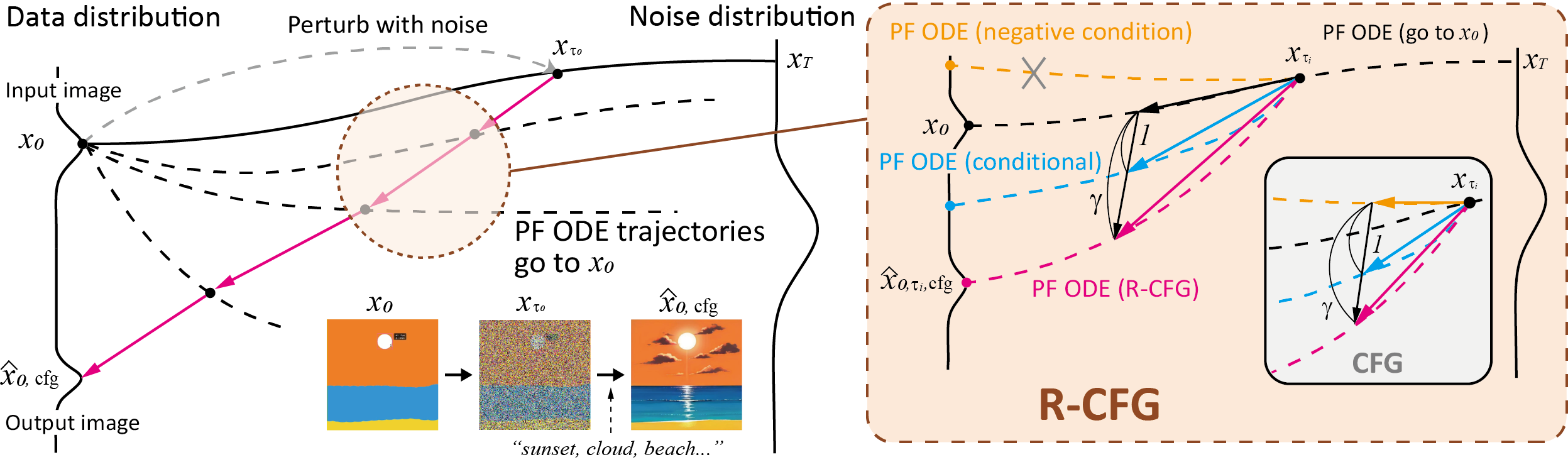}
\caption{Virtual residual noise vectors: The orange vectors depict the virtual residual noise that starts from PF ODE trajectory and points to the original input latent $x_0$}
\label{fig:ICFG}
\end{figure*}

\subsection{Stream Batch: Batching the Denoise Step}
In diffusion models, denoising steps are performed sequentially, resulting in a proportional increase in the processing time of U-Net relative to the number of steps. However, to generate high-fidelity images, it is necessary to increase the number of steps. To resolve this problem in interactive diffusion, we propose a method called Stream Batch.

The Stream Batch technique restructures sequential denoising operations into batched processes, wherein each batch corresponds to a predetermined number of denoising steps, as depicted in Fig. \ref{fig:batch_concept}. The size of each batch is determined by the number of these denoising steps. This approach allows for each batch element to advance one step further in the denoising sequence via a single pass through U-Net. By iteratively applying this method, it is possible to effectively transform input images encoded at timestep \textit{t} into their corresponding image-to-image results at timestep \textit{t+n}, thereby streamlining the denoising procedure.

Stream Batch significantly reduces the need for multiple U-Net inferences. The processing time does not escalate linearly with the number of steps. This technique effectively shifts the trade-off from balancing processing time and generation quality to balancing VRAM capacity and generation quality. With adequate VRAM scaling, this method enables the production of high-quality images within the span of a single U-Net processing cycle, effectively overcoming the constraints imposed by increasing denoising steps.

Waiting and Batching can also increase the throughput of the diffusion pipeline. However, with naive Waiting and Batching (WB), denoising cannot begin immediately on the first input frame, leading to higher latency compared to Stream Batch. We mainly aim for smooth streaming applications. Yet achieving a smooth frame rate with WB requires additional engineering, such as precise inference speed estimation and input-output frame synchronization, and minor timing errors must be carefully managed. In contrast, Stream Batch automatically ensures a consistent interval between input and output frames, providing the advantage of lower latency while dynamically reaching the optimal throughput.

\subsection{Improve Time Consistency by Stream Batch}
Maintaining temporal consistency in video generation is challenging. Many approaches ensure frame coherence by referencing past frames, often through cross-frame attention. However, our Stream Batch method uniquely enables temporal consistency using information from future frames. As shown in Fig. \ref{fig:batch_concept}, Stream Batch allows simultaneous denoising of multiple frames, passing information from future frames to the current frame. This supports real-time image translation that adapts to sudden changes in input while preserving consistency.
In Stream Batch with \( n \) denoising steps, keys and values for each frame at each time step form the following batches:
\[
K_{\text{Batch}} = \left[ K_{t+i,0}, \dots, K_{t,i}, \dots, K_{t-(n-1-i), n-1} \right]
\]
\[
V_{\text{Batch}} = \left[ V_{t+i,0}, \dots, V_{t,i}, \dots, V_{t-(n-1-i), n-1} \right]
\]

These key and value batches incorporate information across different time frames and denoising steps. For example, if the frame at time step \( t \) has reached the \( i \)-th denoising step, the batch includes \( i \) future denoising steps for different frames and \( n-1-i \) past frames. In Stream Batch Cross-frame Attention, rather than using the typical \( K_{t,i} \) and \( V_{t,i} \), we employ \( K_{\text{Batch}} \) and \( V_{\text{Batch}} \), which integrate past and future frame information for the attention computation:

\[
\text{Attn}(Q_{t,i}, K_{\text{Batch}}, V_{\text{Batch}}) = \text{Softmax}\left(\frac{Q_{t,i} \cdot K_{\text{Batch}}^{T}}{\sqrt{d}}\right) V_{\text{Batch}}
\]

This approach enables the generation process to account for information from both past and future frames, as well as across different denoising stages, thus enhancing temporal consistency.

\subsection{Residual Classifier-Free Guidance}

Firstly, SDEdit based method \cite{meng2022sdedit} adds perturbation to the input image $x_0$ and transfers it to the noise distribution $x_{\tau_0}$ as follows,

\begin{equation}
x_{\tau_0} = \sqrt{\alpha_{\tau_0}}x_0 + \sqrt{\beta_{\tau_0}}\epsilon_0,
\label{eq:adding_noise}
\end{equation}
where $\alpha_{\tau_0}$ and $\beta_{\tau_0}$ are values determined by a noise scheduler and $\epsilon_0$ is a sampled noise from a Gaussian $\mathcal{N}(0, I)$. 
When using consistency models for conditional image editing, $x_{\tau_0}$ can be considered as a point on the PF ODE trajectory, which leads to the conditioning manifold. To intensify the conditioning by Classifier-Free Guidance (CFG)\cite{NEURIPS2020_49856ed4}, it is imperative to compute a noise for a negative condition PF ODE trajectory, which is used in vector arithmetic shift for the guidance (Eq.~\ref{eq:cfg}). 

\begin{equation}
\epsilon_{\tau_{i}, \mathrm{cfg}} = \epsilon_{\tau_{i}, \bar{c}} + \gamma(\epsilon_{\tau_{i}, c} - \epsilon_{\tau_{i}, \bar{c}}),
\label{eq:cfg}
\end{equation}
This requirement introduces additional computational overhead at each denoising step.
To reduce this computational overhead, R-CFG utilizes the fact that the original input image $x_0$ is referable at any stage of denoising steps.

For any latent $x_{\tau_i}$ at the denoising step $\tau_i$, we can assume the existence of the virtual negative condition $\bar{c}_{\tau_{i}}^\prime$, that satisfies the self-consistency described as Eq.~\ref{eq:self_consistency}
This implies that $x_{\tau_i}$ is on the PF ODE trajectory going back to the input image $x_0$.

\begin{equation}
x_0 \approx \hat{x}_{0,{\tau_{i}},\bar{c}_{\tau_{i}}^\prime} = f_\theta(x_{{\tau_{i}}}, \tau_{i}, \bar{c}_{\tau_{i}}^\prime)
\label{eq:self_consistency}
\end{equation}

Following the LCM model parameterization \cite{luo2023lcm} and our approximation for the inference time skip connections (\(c_\mathrm{skip}(\tau)=0\), \(c_\mathrm{out}(\tau)=1\) at \(\tau\neq0\)), the self-consistency equation (Eq.~\ref{eq:self_consistency}) can be expressed as follows,

\begin{equation}
x_0 \approx \frac{x_{{\tau_{i}}} - \sqrt{\beta_{\tau_{i}}}\epsilon_{{\tau_{i}},\bar{c}_{\tau_{i}}^\prime}}{\sqrt{\alpha_{\tau_{i}}}}
\label{eq:virtual_x_0_predict}
\end{equation}

Given the initial value $x_0$, and the subsequent values of $x_{{\tau_{i}}}$ obtained sequentially through the iterative denoising, the virtual noise vector $\epsilon_{{\tau_{i}},\bar{c}^\prime}$ in the direction toward the input image can be analytically determined by employing these values with the Eq. \ref{eq:virtual_x_0_predict}:

\begin{equation}
\epsilon_{{\tau_{i}},\bar{c}^\prime} = \frac{x_{{\tau_{i}}} - \sqrt{\alpha_{\tau_{i}}}x_0}{\sqrt{\beta_{\tau_{i}}}}
\label{eq:virtual_residual_noise_from_x0}
\end{equation}

With the virtual noise $\epsilon_{{\tau_{i}},\bar{c}^\prime}$ obtained from Eq.~\ref{eq:virtual_residual_noise_from_x0}, we formulate R-CFG by:
\begin{equation}
\epsilon_{\tau_{i}, \mathrm{cfg}} = \delta\epsilon_{\tau_{i}, \bar{c}^\prime} + \gamma(\epsilon_{\tau_{i}, c} - \delta\epsilon_{\tau_{i}, \bar{c}^\prime})
\label{eq:next_step_cfg}
\end{equation}
where $\delta$ is a magnitude moderation coefficient for the virtual residual noise that softens the effect and the approximation error of the virtual residual noise.

R-CFG that uses the original input image latent $x_0$ as the residual term can effectively generate results that diverge from the original input image according to the magnitude of the guidance scale $\gamma$, thereby enhancing the effect of conditioning without the need for additional U-Net computations. We call this method Self-Negative R-CFG.

Not only to deviate from the original input image $x_0$, but also to diverge from any negative condition, we can find the desired reference point $\hat{x}_{0,{\tau_{0}},\bar{c}}$ that reflects the negative condition $\bar{c}$ using the same self-consistency formulation:

\begin{equation}
\hat{x}_{0,{\tau_{0}},\bar{c}} = \frac{x_{\tau_{0}} - \sqrt{\beta_{\tau_{0}}}\epsilon_{{\tau_{0}},\bar{c}}}{\sqrt{\alpha_{\tau_{0}}}}
\label{eq:negative_conditioned_x_0_predict}
\end{equation}

We can obtain $\hat{x}_{0,{\tau_{0}},\bar{c}}$ by computing the actual negative conditioned noise $\epsilon_{\tau_{0},\bar{c}}$ using U-Net only one time for the first denoising step.

In Eq. \ref{eq:virtual_residual_noise_from_x0}, instead of $x_0$, using $\hat{x}_{0,{\tau_{0}},\bar{c}}$, we can obtain the virtual negative conditioned noise $\epsilon_{{\tau_{i+1}},\bar{c}^\prime}$ that can effectively diverge the generation results from the controllable negative conditioning $\bar{c}$. We name this method Onetime-Negative R-CFG.
In contrast to the conventional CFG, which requires $2n$ computations of U-Net, the Self-Negative RCFG and Onetime-Negative RCFG necessitate only $n$ and $n+1$ computations of U-Net, respectively, where $n$ is the number of the denoising steps.

\subsection{Stochastic Similarity Filter}
When images remain unchanged or show minimal changes, particularly in scenarios without active user interaction or static environment, nearly identical input images are often repeatedly fed into the VAE and U-Net. This leads to the generation of identical or nearly identical images and unnecessary consumption of GPU resources. In contexts involving continuous inputs, such instances of unmodified input images can occasionally occur. To tackle this issue and minimize unnecessary computational load, we propose a strategy termed \textit{stochastic similarity filter (SSF)}, as shown in Fig. \ref{fig:overview}.

We calculate the cosine similarity between the current input image $I_t$ and the past reference frame image $I_\mathrm{ref}$.

\begin{equation}
S_{C}(I_t, I_\mathrm{ref}) = \frac{I_t \cdot I_\mathrm{ref}}{\|I_t\|\|I_\mathrm{ref}\|}
\label{eq:cos-simg}
\end{equation}

Based on this cosine similarity, we calculate the probability of skipping the subsequent VAE and U-Net processes. It is given by

\begin{equation}
\mathbf{P}(\mathrm{skip} | I_t, I_\mathrm{ref}) = \mathbf{max}\left\{0,\: \frac{S_{C}(I_t, I_\mathrm{ref}) - \eta}{1-\eta}\right\},
\label{eq:skip_prob}
\end{equation}
where $\eta$ is the similarity threshold. This probability decides whether subsequent processes like VAE Encoding, U-Net, and VAE Decoding should be skipped or not. If not skipped, the input image at that time is saved and updated as the reference image $I_{ref}$ for future use. This probabilistic skipping mechanism allows the network to operate fully in dynamic scenes with low inter-frame similarity, while in static scenes with high inter-frame similarity, the network's operational rate decreases, conserving computational resources. The GPU usage is modulated seamlessly based on the similarity of the input images, enabling smooth adaptation to scenes with varying dynamics. 

\begin{figure*}[!t]
  \centering
  \begin{minipage}{0.48\linewidth}
    \includegraphics[width=\linewidth]{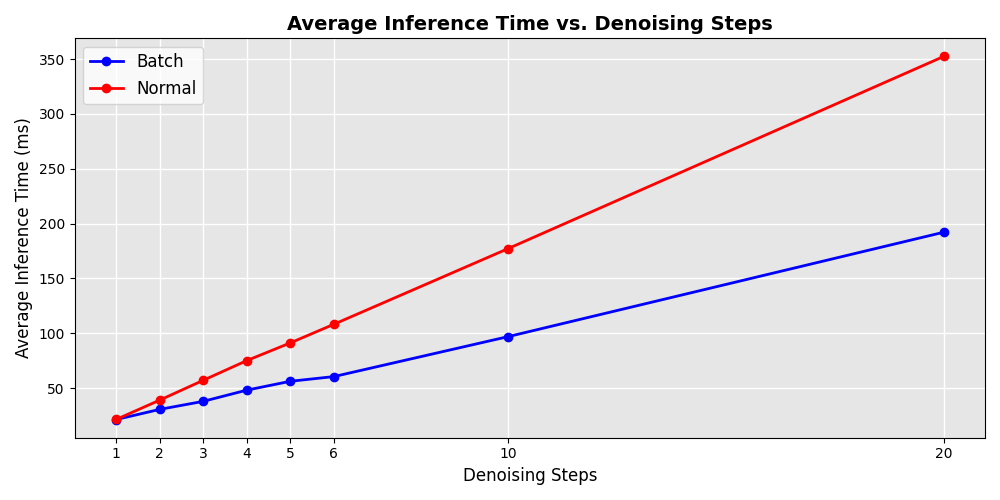}
    \caption{Average inference time comparison between Stream Batch and normal sequential denoising without TensorRT.}
    \label{fig:infe_comp_withoutTRT}
  \end{minipage}
  \hfill 
  \begin{minipage}{0.48\linewidth}
    \includegraphics[width=\linewidth]{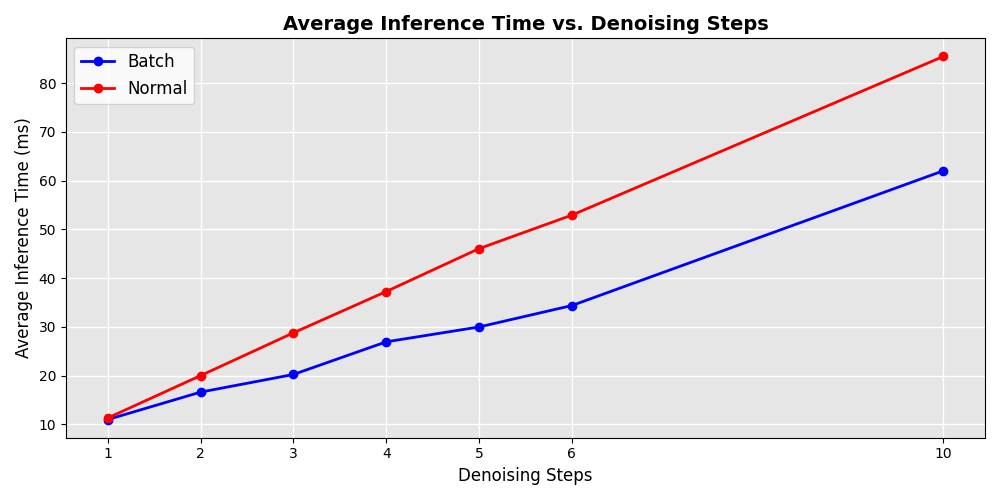}
    \caption{Average inference time comparison between Stream Batch and normal sequential denoising with using TensorRT.}
    \label{fig:infe_comp_withTRT}
  \end{minipage}
\end{figure*}

\textit{Note:} We emphasize that compared to determining whether we skip the compute via a hard threshold, the proposed probability-sampling-based similarity filtering strategy leads to a smoother video generation. Because the hard threshold is prone to making the video stuck, which hurts the impression of watching video streaming, while the sampling-based method significantly improves the smoothness. For the other efficiency improvement methods, we illustrate them in the supplementary material.

\section{Experiments}
\label{sec:experiments}

We implement StreamDiffusion pipeline upon LCM, LCM-LoRA \cite{luo2023lcm, luo2023lcmlora} and SD-turbo \cite{sauer2023adversarial}. As a model accelerator, we use TensorRT and for the lightweight efficient VAE, we use TAESD \cite{kingma2022autoencoding}. Our pipeline is compatible with the customer-level GPU. We test our pipeline on NVIDIA RTX4090 GPU, Intel Core i9-13900K CPU, Ubuntu 22.04.3 LTS, and NVIDIA RTX3060 GPU, Intel Core i7-12700K, Windows 11 for image generation. We note that we evaluate the throughput mainly via the average inference time per image through processing 100 images.

\subsection{Quantitative Evaluation}
We compare our method with the AutoPipelineForImage2Image, which is a pipeline developed by Huggingface diffusers \footnote{\url{https://github.com/huggingface/diffusers}}. The average inference time comparison is presented in Table. \ref{tb:pipeline_comparison}. Our pipeline demonstrates a substantial speed increase. When we use TensorRT, StreamDiffusion achieves a minimum speed-up of 13.0 times when running the 10 denoising steps, and reaching up to 59.6 times in scenarios involving a single denoising step. Even though without TensorRT, StreamDiffusion achieves a 29.7 times speed up compared to AutoPipeline when using one step denoising, and an 8.3 times speedup at 10 step denoising.

\begin{table*}[!htpb]
\centering
\caption{Comparison of average inference time (ms) at different denoising steps with speedup factors. The first column denotes the denoising steps and the AutoPipeline is from Diffusers \cite{diffusers}. }
\label{tb:pipeline_comparison}
\footnotesize 
\begin{tabular}{c|c|c|c}
\toprule
\textbf{Step} & \textbf{StreamDiffusion} & \textbf{StreamDiffusion w/o TRT} & \textbf{AutoPipeline Img2Img} \\
\midrule
1 & 10.65 (59.6x) & 21.34 (29.7x) & 634.40 (1x) \\
2 & 16.74 (39.3x) & 30.61 (21.3x) & 652.66 (1x) \\
4 & 26.93 (25.8x) & 48.15 (14.4x) & 695.20 (1x) \\
10 & 62.00 (13.0x) & 96.94 (8.3x) & 803.23 (1x) \\
\bottomrule
\end{tabular}
\end{table*}

\paragraph{\textbf{Efficiency comparison regarding Stream Batch.}}
The efficiency comparison between Stream Batch and the original sequential U-Net loop is shown in Fig. \ref{fig:infe_comp_withoutTRT}.
When implementing a denoising batch strategy, we observe a significant improvement in processing time. It achieves a reduction by half when compared to a conventional U-Net loop at sequential denoising steps. Even though applying TensorRT, the accelerator tool for neural modules, our proposed Stream Batch still boosts the efficiency of the original sequential diffusion pipeline by a large margin at different denoising steps.

\paragraph{\textbf{Efficiency comparison regarding R-CFG.}}
Table. \ref{tb:cfg_comparision} presents a comparison of the inference times for StreamDiffusion pipelines with R-CFG and conventional CFG. The additional computations required to apply Self-Negative R-CFG are merely lightweight vector operations, resulting in negligible changes in inference time compared to when Self-Negative is not used. When employing Onetime-Negative R-CFG, additional UNet computations are necessary for the first step of the denoising process. Therefore, One-time-negative R-CFG and conventional CFG have almost identical inference times for a single denoising step case. However, as the number of denoising steps increases, the difference in inference time from conventional CFG to both Self-Negative and Onetime-Negative R-CFG becomes more pronounced. At denoising step 5, a speed improvement of 2.05x is observed with Self-Negative R-CFG and 1.79x with Onetime-Negative R-CFG, compared to conventional CFG.

\begin{table*}[!t]
\centering
\caption{Comparison of average inference time (ms) at different denoising steps among different CFG methods}
\label{tb:cfg_comparision}
\begin{tabular}{c|c|c|c}
\toprule
\textbf{Step} & \textbf{Self-Negative R-CFG} & \textbf{Onetime-Negative R-CFG} & \textbf{CFG} \\
\midrule
1 & 11.04 (1.52x) & 16.55 (1.01x) & 16.74 (1x) \\
2 & 16.61 (1.64x) & 20.64 (1.32x) & 27.18 (1x) \\
3 & 20.64 (1.74x) & 27.25 (1.32x) & 35.91 (1x) \\
4 & 26.19 (1.90x) & 31.65 (1.57x) & 49.71 (1x) \\
5 & 31.47 (2.05x) & 36.04 (1.79x) & 64.64 (1x) \\
\bottomrule
\end{tabular}
\end{table*}

\subsection{Energy Consumption}

We then conduct a comprehensive evaluation of the energy consumption associated with our proposed stochastic similarity filter (SSF), as depicted in Figure. \ref{fig:gpu_usage_3090} and Figure. \ref{fig:gpu_usage_4090}. These figures provide the GPU utilization patterns when SSF (Threshold $\eta$ set at 0.98) is applied to input videos containing scenes with periodic static characteristics. The comparative analysis reveals that the incorporation of SSF significantly mitigates GPU usage in instances where the input images are predominantly static and demonstrate a high degree of similarity.

Figure. \ref{fig:gpu_usage_3090} delineates the results derived from a meticulously executed two-denoise-step img2img experiment. This experiment was conducted on a 20-frame video sequence, employing NVIDIA RTX3060 graphics processing units with or without the integration of SSF. The experiment results indicate a substantial decrease in average power consumption from \textbf{85.96w} to \textbf{35.91w} on one RTX3060 GPU. Using the same static scene input video with one NVIDIA RTX4090GPU, the power consumption was reduced from  \textbf{238.68w} to \textbf{119.77w}.

Furthermore, Figure. \ref{fig:gpu_usage_4090} expounds on the findings from a similar two-denoise-step img2img experiment using one RTX4090GPU. This time the evaluation of energy consumption is performed on a 1000-frame video featuring dynamic scenes. Remarkably, even under drastically dynamic conditions, the SSF efficiently extracted several frames exhibiting similarity from the dynamic sequence. This process results in a noteworthy reduction in average power consumption, from \textbf{236.13w} to \textbf{199.38w}. These findings underscore the efficacy of the Stochastic Similarity Filter in enhancing energy efficiency, particularly in scenarios involving static or minimally varying visual content.


\subsection{Ablation study}

\begin{figure}[!htp]
  \centering
   \includegraphics[width=1.0\linewidth]{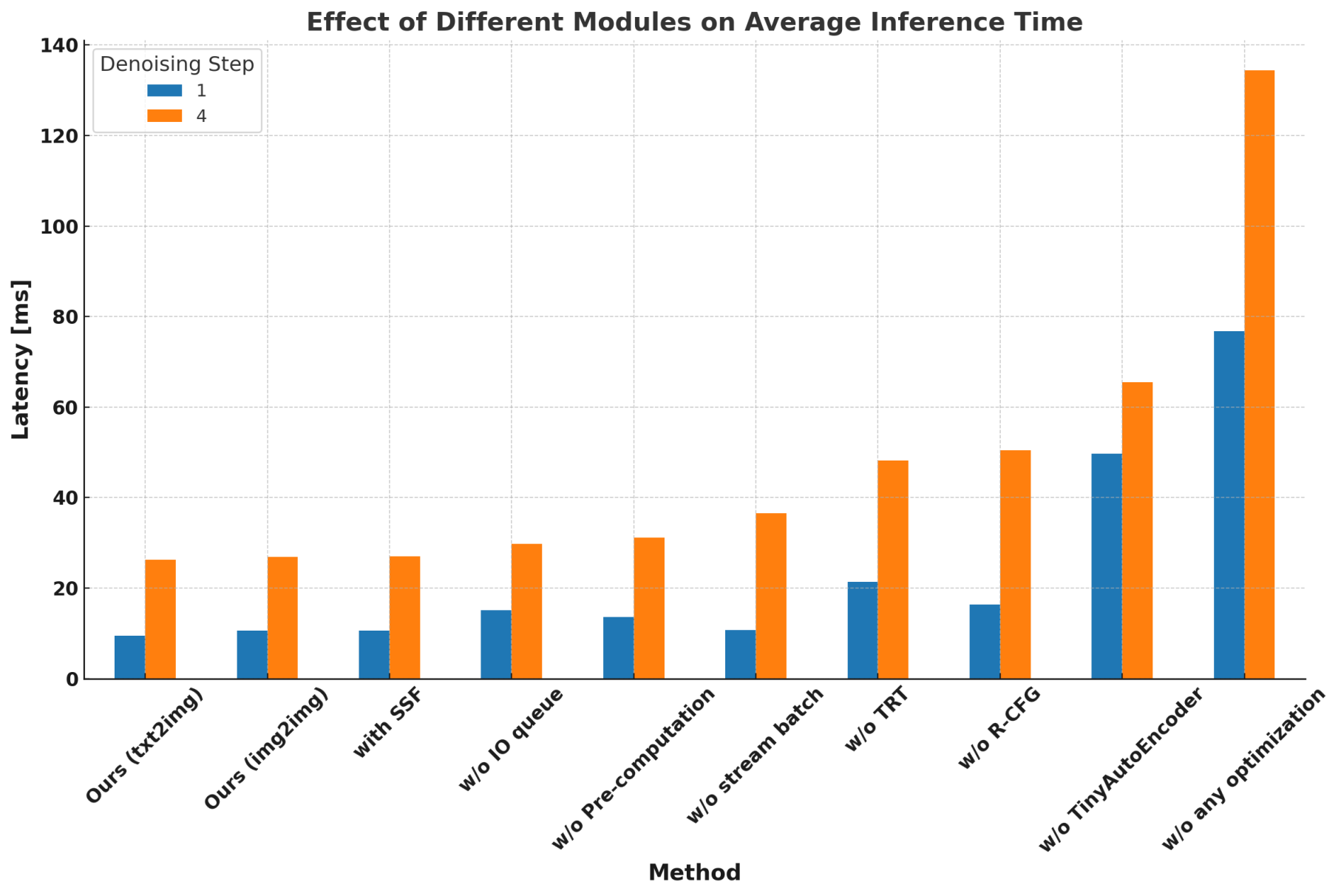}
    \caption{Ablation study on different components. }
\label{fig:diff-compo}
\end{figure}

In our ablation study, as summarized in Fig. \ref{fig:diff-compo}, we evaluate the average inference time of our proposed method under various configurations to understand the contribution of each component. Our proposed StreamDiffusion achieves an average inference time of 10.98/9.42 ms and 26.93/26.30 ms for denoising steps 1 and 4 on image-to-image/text-to-image generation, respectively. When the R-CFG is not used, we observe this results in the second largest efficiency drop. This demonstrates that R-CFG is one of the most critical components in our pipeline. When the Stream Batch processing is removed ('w/o stream batch'), we observe a large time consumption increase, especially at 4 denoising steps. We also evaluate the impact on the inference time of our pipeline regarding SSF. We observe that SSF plays a significant role in enabling energy saving and does not introduce extra time cost.

Besides, the absence of TensorRT ('w/o TRT') leads to a large increase in time cost. The removal of pre-computation also results in increased time cost but not much. We attribute the reason to the limited number of key-value computations in Stable Diffusion. Besides, the exclusion of input-output queue ('w/o IO queue') also demonstrates an impact on average inference time, which mainly aims to optimize the parallelization issue resulting from pre- and post-processing. 
In the AutoPipelineImage2Image's adding noise function, the precision of tensors is converted from fp32 to fp16 for each request, leading to a decrease in speed. In contrast, the StreamDiffusion pipeline standardizes the precision of variables and computational devices beforehand. It does not perform tensor precision conversion or computation device transfers during inference. Consequently, even without any optimization ('w/o any optimization'), our pipeline significantly outperforms the AutoPipelineImage2Image in terms of speed.

\subsection{Quantitative Evaluation for the Image Quality.}
We conduct a quantitative evaluation on the image quality. Specifically, we first evaluate the FID and CLIP scores on the text-to-image generation. We use the same dataset as \cite{brooks2022instructpix2pix} for the evaluation. We use LCM as our main baseline for the comparison. Note that our method is never trained; our method still improves LCM by a large margin in terms of the FID (29.69 vs. 26.79) and maintains a similar CLIP score (24.95 vs. 24.99). This demonstrates the effectiveness of our proposed method.

\paragraph{\textbf{User study.}}
We also conduct a user study to validate the visual quality of different components of our StreamDiffusion.
The results are shown in the Table. \ref{tab:combined_user_study}. Our Stream Batch with future-frame attention significantly enhances time consistency compared to its absence. Additionally, our SSF method addresses the crucial yet rarely explored issue of energy efficiency. While SSF is simple, it is far from trivial: a common approach might apply a hard threshold on similarity to regulate streaming flow. However, as noted in the main text, we introduce a novel approach—using probability sampling to achieve superior streaming quality. As Table \ref{tab:combined_user_study} illustrates, this approach is preferred by more users over the hard-threshold method for visual quality. Moreover, compared to the vanilla CFG method, both our self-Negative R-CFG and one-time R-CFG are more preferred by the users, demonstrating our method not only improves efficiency but also visual quality.

\begin{table}[h]
\centering
\resizebox{0.48\textwidth}{!}{
\begin{tabular}{c|cc}
\toprule
\textbf{\# Users} & \textbf{Without Future-Frame (\%)} & \textbf{With Future-Frame (\%)} \\
\midrule
486 & 9.67 & 90.33 \\
\midrule
\textbf{\# Users} & \textbf{Without SSF (\%)} & \textbf{With SSF (\%)} \\
\midrule
144 & 48.61 & 51.39 \\
\midrule
\textbf{\# Users} & \textbf{Hard Threshold (\%)} & \textbf{SSF (\%)} \\
\midrule
45 & 24.44 & 75.56 \\
\midrule
\textbf{\# Users} & \textbf{With / Without CFG (\%)} & \textbf{Self-Neg R-CFG / One-Time R-CFG (\%)} \\
\midrule
257 & 22 / 20 & 35 / 23 \\
\bottomrule
\end{tabular}
}
\vspace{-0.2cm}
\caption{User study results: Future-frame attention consistency (486 users), SSF quality imperceptibility (144 users), streaming quality between hard threshold filter and SSF (45 users), comparison of with/without CFG and self-negative/one-time R-CFG (257 users).}
\label{tab:combined_user_study}
\end{table}

This pipeline enables image generation with very low throughput from input images received in real-time from cameras or screen capture devices. At the same time, it is capable of producing high-quality images that effectively align to the specified prompt conditions. These capabilities demonstrate the applicability of our pipeline in various real-time applications, such as real-time game graphic rendering, generative camera effect filters, real-time face conversion, and AI-assisted drawing.

The alignment of generated images to prompt conditioning using Residual Classifier-Free Guidance (R-CFG) is depicted in Fig. \ref{fig:cfg_conparision}. The generated images, without using any form of CFG, exhibit weak alignment to the prompt, particularly in aspects like color changes or the addition of non-existent elements, which are not effectively implemented. In contrast, the use of CFG or R-CFG enhances the ability to modify original images, such as changing hair color, adding body patterns, and even incorporating objects like glasses. Notably, the use of R-CFG results in a stronger influence of the prompt compared to standard CFG. R-CFG, although limited to image-to-image applications, can compute the vector for negative conditioning while continuously referencing the latent value of the input image and the initially sampled noise. This approach yields more consistent directions for the negative conditioning vector compared to the standard CFG, which uses UNet at every denoising step to calculate the negative conditioning vector. Consequently, this leads to more pronounced changes from the original image. However, there is a trade-off in terms of the stability of the generated results. While Self-Negative R-CFG enhances the prompt's effectiveness, it also has the drawback of increasing the contrast of the generated images. To address this, adjusting the $delta$ in Eq. \ref{eq:next_step_cfg} can modulate the magnitude of the virtual residual noise vector, thereby mitigating the rise in contrast. 
Additionally, using Onetime-Negative R-CFG with appropriately chosen negative prompts can mitigate contrast increases while improving prompt adherence, as observed in Fig. \ref{fig:cfg_conparision}. This approach allows the generated images to blend more naturally with the original image.

Besides, Fig. \ref{fig:time_vis} in appendix shows the image-to-image generation results using StreamBatch Cross-frame attention, with 4 denoising steps. As evident from the figure, compared to the results of StreamDiffusion without Cross-frame attention, the method incorporating information from future and past frames exhibits increased temporal consistency.

\section{Conclusion}

We propose StreamDiffusion, a pipeline-level solution for interactive diffusion generation. StreamDiffusion consists of several optimization strategies for both throughput and GPU usage, including Stream Batch with cross-attention, residual classifier-free guidance (R-CFG), IO-queue for parallelization, stochastic similarity filter, pre-computation, Tiny AutoEncoder, and the use of the model acceleration tool. The synergistic combination of these elements results in a marked improvement in efficiency. Specifically, StreamDiffusion achieves up to 91.07 frames per second (fps) on a standard consumer-grade GPU for image generation tasks. This performance level is particularly beneficial for a variety of applications, including but not limited to the Metaverse, online video streaming, and broadcasting sectors. Furthermore, StreamDiffusion demonstrates a significant reduction in GPU power consumption, achieving at least a 1.99x decrease. This notable efficiency gain underscores StreamDiffusion's potential for commercial application, offering a compelling solution for energy-conscious, high-performance computing environments. 
\section{Acknowledgments}
We sincerely thank Taku Fujimoto and Huggingface team for their invaluable feedback, courteous support, and insightful discussions.
{
    \small
    \bibliographystyle{ieeenat_fullname}
    \bibliography{main}
}
\clearpage
\appendix

\subsection{Qualitative Results}
\begin{figure*}[!t]
  \centering
   \includegraphics[width=.94 \linewidth]{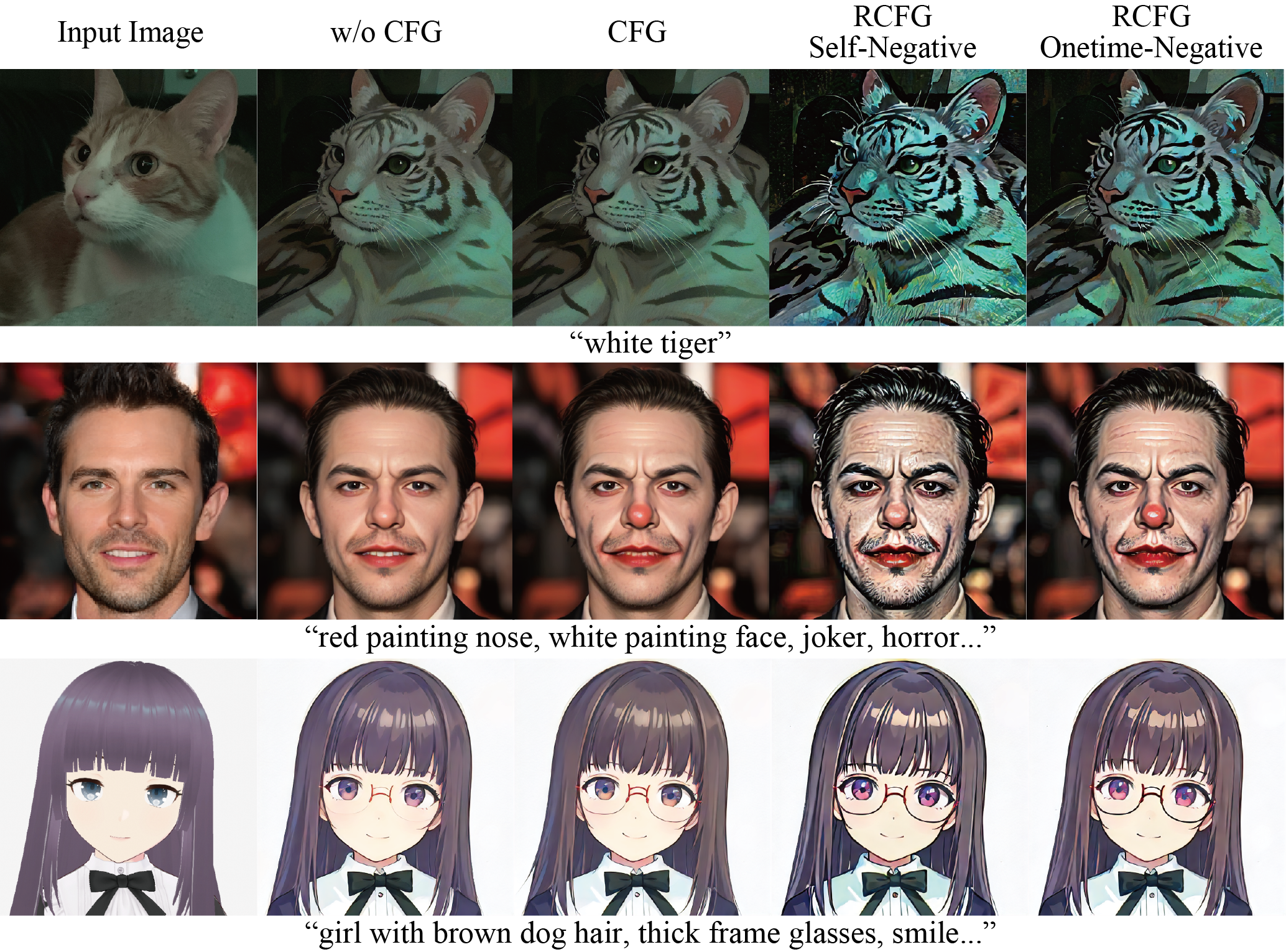}
    \caption{Results using no CFG, standard CFG, and R-CFG with Self-Negative and Onetime-Negative approaches. When compared to cases where CFG is not utilized, the cases with CFG utilized can intensify the impact of prompts. In the proposed method R-CFG, a more pronounced influence of prompts was observed. Both CFG and R-CFG use guidance scale $\gamma = 1.4$. For R-CFG, the first two rows use magnitude modelation coefficient $\delta = 1.0$, and the third row uses $\delta=0.5$.}
\label{fig:cfg_conparision}
\end{figure*}

\begin{figure*}[!t]
  \centering
   \includegraphics[width=\linewidth]{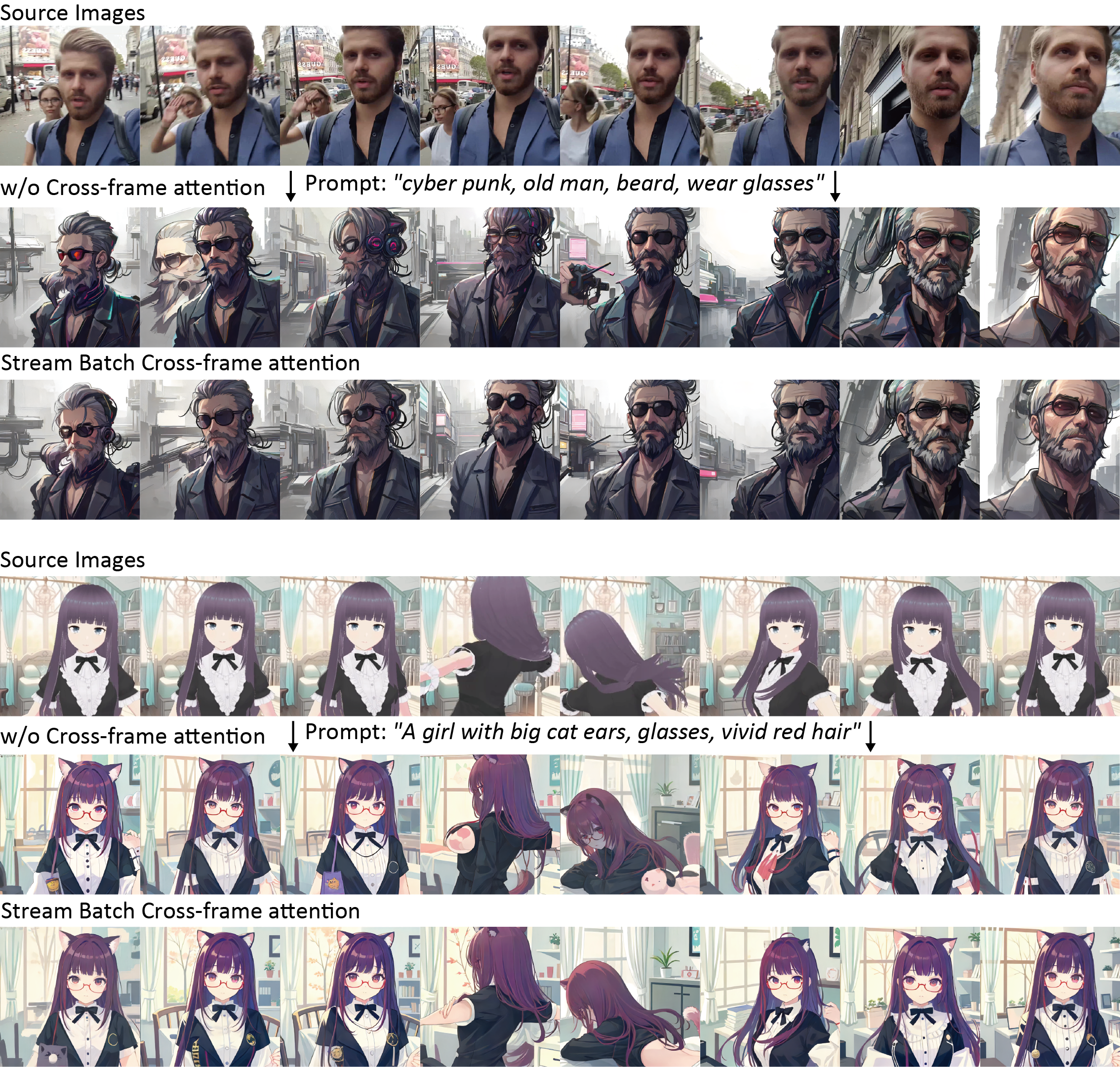}
\caption{Time consistency qualitative evaluation: In cases where the subject's face moves significantly in intermediate frames, it can be observed that using StreamBatch Cross-frame attention produces more appropriate and temporally consistent generation results by leveraging the context from preceding and succeeding frames.}
\label{fig:time_vis}
\end{figure*}

\section{More Architecture details}

\subsection{Input-Output Queue}

\begin{figure*}[!t]
  \centering
   \includegraphics[width=.9\linewidth]{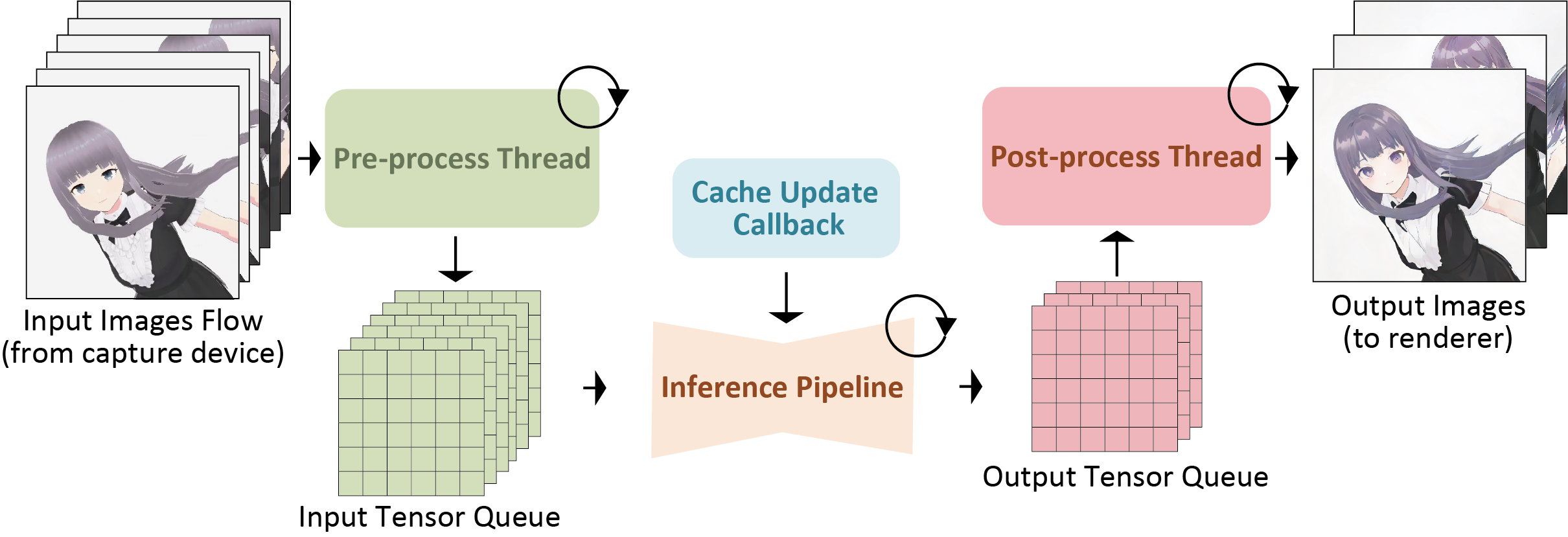}
\caption{Input-Output Queue: The process of converting input images into a tensor data format manageable by the pipeline, and conversely, converting decoded tensors back into output images requires a non-negligible amount of additional processing time. To avoid adding these image processing times to the bottleneck process, the neural network inference process, we have segregated image pre-processing and post-processing into separate threads, allowing for parallel processing. Moreover, by utilizing an Input Tensor Queue, we can accommodate temporary lapses in input images due to device malfunctions or communication errors, enabling smooth streaming.}
\label{fig:queue_para}
\end{figure*}

The current bottleneck in high-speed image generation systems lies in the neural network modules, including VAE and U-Net. To maximize the overall system speed, processes such as pre-processing and post-processing of images, which do not require handling by the neural network modules, are moved outside of the pipeline and processed in parallel.

In the context of input image handling, specific operations, including resizing of input images, conversion to tensor format, and normalization, are meticulously executed. To address the disparity in processing frequencies between the human inputs and the model throughput, we design an input-output queuing system to enable efficient parallelization, as shown in Fig. \ref{fig:queue_para}. This system operates as follows: processed input tensors are methodically queued for Diffusion Models. During each frame, Diffusion Model retrieves the most recent tensor from the input queue and forwards it to the VAE Encoder, thereby triggering the image generation sequence. Correspondingly, tensor outputs from the VAE Decoder are fed into an output queue. In the subsequent output image handling phase, these tensors are subject to a series of post-processing steps and conversion into the appropriate output format. Finally, the fully processed image data is transmitted from the output handling system to the rendering client.

\subsection{Pre-computation}
The U-Net architecture requires both input latent variables and conditioning embeddings. Typically, the conditioning embedding is derived from a text prompt, which remains constant across different frames. To optimize this, we pre-compute the prompt embedding and store it in a cache. In interactive or streaming mode, this pre-computed prompt embedding cache is recalled. 
Within U-Net, the Key and Value are computed based on this pre-computed prompt embedding for each frame. We have modified the U-Net to store these Key and Value pairs, allowing them to be reused. Whenever the input prompt is updated, we recompute and update these Key and Value pairs inside U-Net.

For consistent input frames across different timesteps and to improve computational efficiency, we pre-sample Gaussian noise for each denoising step and store it in the cache. This approach is particularly relevant for image-to-image tasks.

We also precompute \(\alpha_{\tau}\) and \(\beta_{\tau}\), the noise strength coefficients for each denoising step \(\tau\), defined as:

\begin{equation}
x_t = \sqrt{\alpha_{\tau}}x_0 + \sqrt{\beta_{\tau}}\epsilon
\end{equation}
\label{eq:fw_diffusion_process}

This is a minor point in low throughput scenarios, but at frame rates higher than 60 FPS, the overhead of recomputing these static values becomes noticeable.

We note that we have a specific design for the inference parameterization for latent consistency models (LCM). As per the original paper, we need to compute \(c_\mathrm{skip}(\tau)\) and \(c_\mathrm{out}(\tau)\) to satisfy the following equation:

\begin{equation}
f_\theta(x,\tau) = c_\mathrm{skip}(\tau)x + c_\mathrm{out}(\tau)F_\theta(x,\tau).
\end{equation}
\label{eq:cm_parameterization}

The functions \(c_\mathrm{skip}(\tau)\) and \(c_\mathrm{out}(\tau)\) in original LCM \cite{luo2023lcm} is constructed as follows:

\begin{equation}
c_\mathrm{skip}(\tau)=\frac{\sigma_\mathrm{data}^2}{(s\tau)^2+\sigma_\mathrm{data}^2}, \quad c_\mathrm{out}(\tau)=\frac{\sigma_\mathrm{data}s\tau}{\sqrt{\sigma_\mathrm{data}^2+(s\tau)^2}},
\end{equation}
\label{eq:c_skip_c_out}

where \(\sigma_{\mathrm{data}}=0.5\), and the timestep scaling factor \(s=10\). We note that with \(s=10\), \(c_\mathrm{skip}(\tau)\) and \(c_\mathrm{out}(\tau)\) approximate delta functions that enforce the boundary condition to the consistency models. (i.e., at denoising step \(\tau=0\), \(c_\mathrm{skip}(0)=1\), \(c_\mathrm{out}(0)=0\); and at \(\tau\neq0\), \(c_\mathrm{skip}(\tau)=0\), \(c_\mathrm{out}(\tau)=1\)). At inference time, there's no need to recompute these functions repeatedly. We can either pre-compute \(c_\mathrm{skip}(\tau)\) and \(c_\mathrm{out}(\tau)\) for all denoising steps \(\tau\) in advance or simply use constant values \(c_\mathrm{skip}=0\), \(c_\mathrm{out}=1\) for any arbitrary denoising step \(\tau\).

\subsection{Model Acceleration and Tiny AutoEncoder}
We employ TensorRT to construct the U-Net and VAE engines, further accelerating the inference speed. TensorRT is an optimization toolkit from NVIDIA that facilitates high-performance deep learning inference. It achieves this by performing several optimizations on neural networks, including layer fusion, precision calibration, kernel auto-tuning, dynamic tensor memory, and more. These optimizations are designed to increase throughput and efficiency for deep learning applications.

To optimize speed, we configured the system to use static batch sizes and fixed input dimensions (height and width). This approach ensures that the computational graph and memory allocation are optimized for a specific input size, leading to faster processing times. However, this means that if there is a requirement to process images with different shapes (i.e., varying heights and widths) or to use different batch sizes (including those for denoising steps), a new engine tailored to these specific dimensions must be built. This is because the optimizations and configurations applied in TensorRT are specific to the initially defined dimensions and batch size, and changing these parameters would necessitate a reconfiguration and re-optimization of the network within TensorRT.

Besides, we employ a tiny AutoEncoder, which has been engineered as a streamlined and efficient counterpart to the traditional Stable Diffusion AutoEncoder \cite{kingma2022autoencoding, rombach2021highresolution}. TAESD excels in rapidly converting latents into full-size images and accomplishing decoding processes with significantly reduced computational demands.

\section{Text-to-Image Quality}
\begin{figure*}[!t]
  \centering
   \includegraphics[width=.9\linewidth]{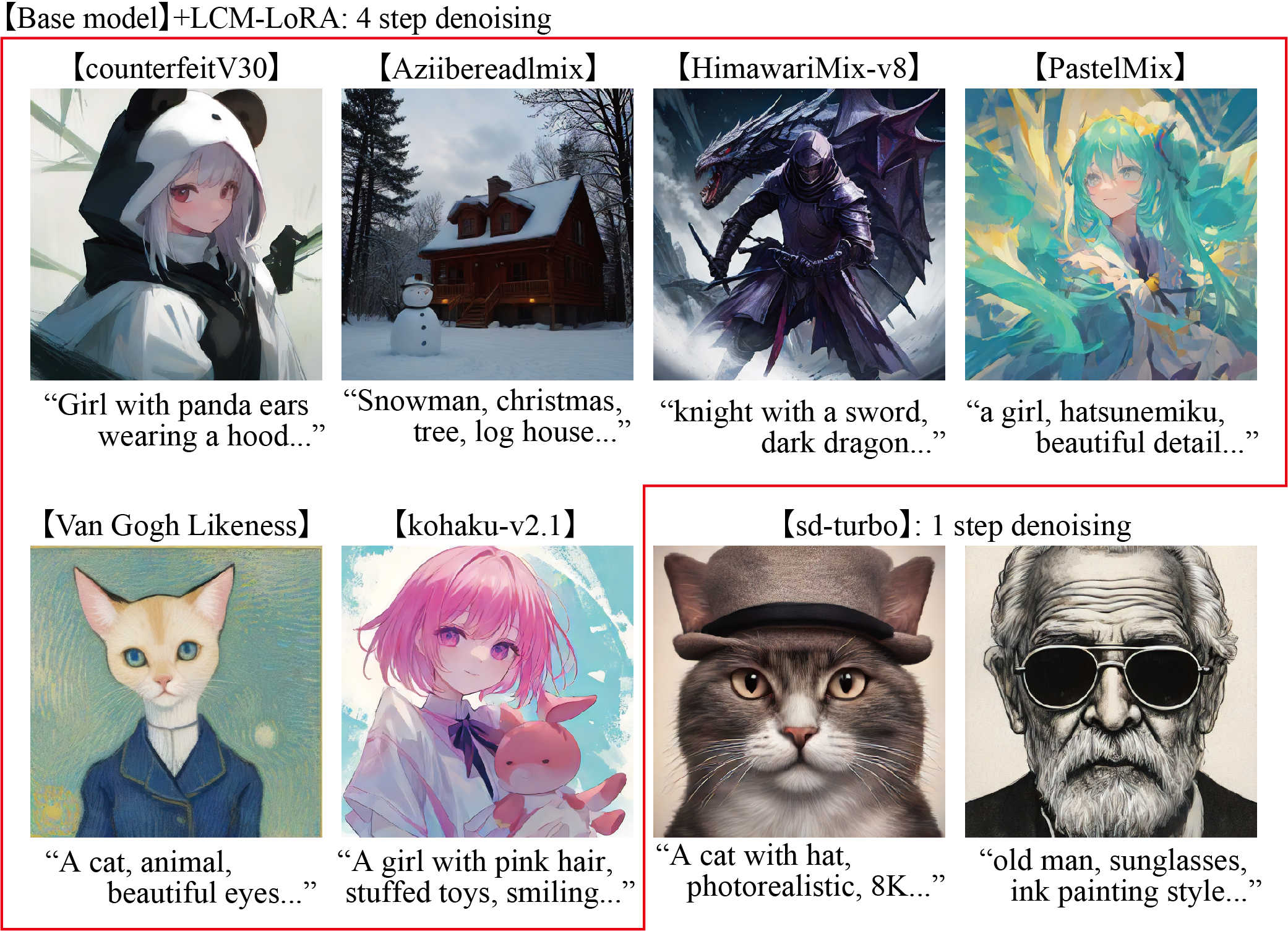}
\caption{Text-to-Image generation results. We use four step denoising for LCM-LoRA, and one step denoising for sd-turbo. Our StreamDiffusion enables the real-time generation of images with quality comparable to those produced using Diffusers AutoPipeline Text2Image.}
\label{fig:txt2img_results}
\end{figure*}

The quality of standard text-to-image generation results is demonstrated in Fig. \ref{fig:txt2img_results}. Using the sd-turbo model, high-quality images like those shown in Fig. \ref{fig:txt2img_results} can be generated in just one step. When images are produced using our proposed StreamDiffusion pipeline and SD-turbo model in an environment with GPU: RTX 4090, CPU: Core i9-13900K, and OS: Ubuntu 22.04.3 LTS, it's feasible to generate such high-quality images at a rate exceeding 100fps. Furthermore, by increasing the batch size of images generated at once to 12, our pipeline can continuously produce approximately 150 images per second.
The images enclosed in red frames shown in Fig. \ref{fig:txt2img_results} are generated in four steps using community models merged with LCM-LoRA. While these LCM models require more than 1 step for high quality image generation, resulting in a reduction of speed to around 40fps, these LCM-LoRA based models offer the flexibility of utilizing any base model, enabling the generation of images with diverse expressions.

\section{GPU Usage Under Dynamic Scene}

\begin{figure*}[!t]
  \centering
   \includegraphics[width=.9\linewidth]{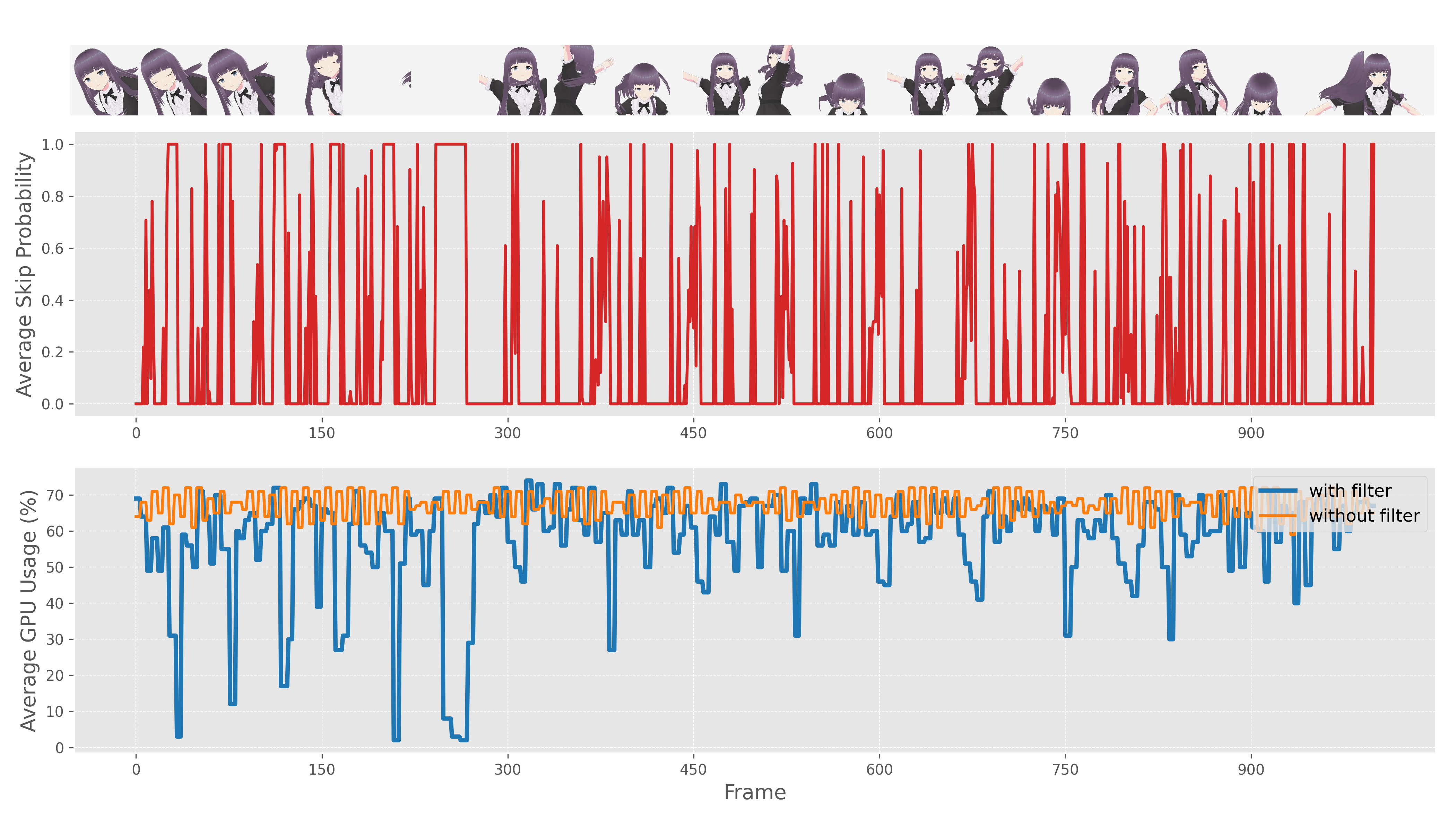}
\caption{GPU Usage comparison under dynamic scene. (GPU: RTX4090, Number of frames: 1000) The blue line represents the GPU usage with SSF, the orange line indicates GPU usage without SSF, and the red line denotes the Skip probability calculated based on the cosine similarity between input frames. Additionally, the top of the plot displays input images corresponding to the same timestamps. In this case, the character in the input images keeps moving dynamically. Thus, this analysis compares GPU usage in a dynamic scenario.}
\label{fig:gpu_usage_4090}
\end{figure*}

We also evaluate the GPU usage under dynamic scenes on one RTX 4090 GPU, as shown in the Figure. \ref{fig:gpu_usage_4090}. The analysis of the GPU usage is shown in Section 4.2 of the main text. 
\begin{figure*}[!t]
    \centering
    \includegraphics[width=.9\linewidth]{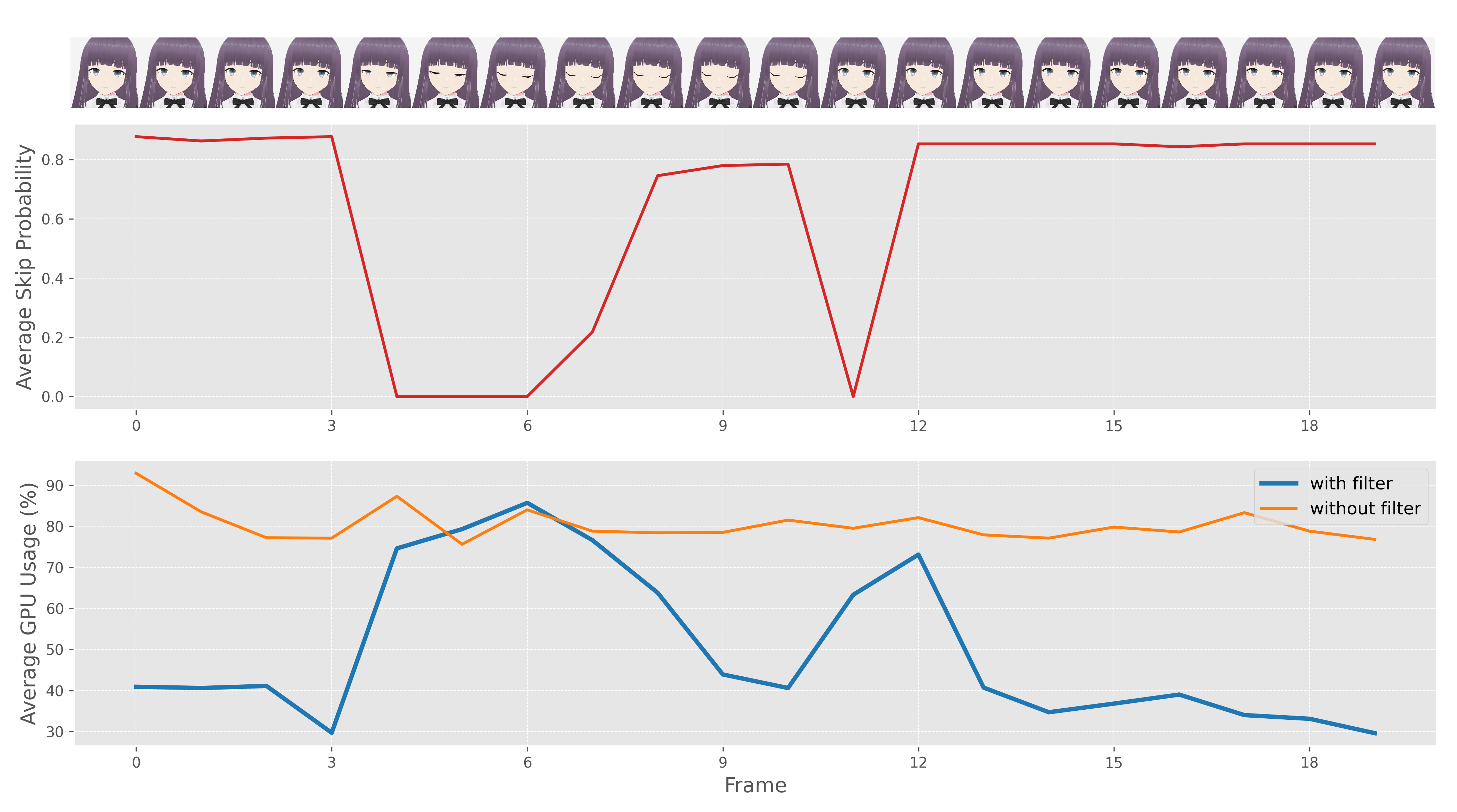}
    \vspace{-5mm}
    \caption{GPU Usage comparison under static scene. (GPU: RTX3060, Number of frames: 20) The blue line represents the GPU usage with SSF, the orange line indicates GPU usage without SSF, and the red line denotes the Skip probability calculated based on the cosine similarity between input frames. Additionally, the top of the plot displays input images corresponding to the same timestamps. In this case, the character in the input images is only blinking.}
    \label{fig:gpu_usage_3090}
    \vspace{-10mm}
\end{figure*}

\end{document}